\documentclass[11pt,letterpaper]{article}

\usepackage[margin=1in]{geometry}
\usepackage{microtype}
\usepackage[T1]{fontenc}
\usepackage[utf8]{inputenc}
\usepackage{lmodern}
\usepackage{textcomp}
\usepackage{amsmath,amssymb}
\usepackage{booktabs,array}
\usepackage{caption}
\usepackage{listings}
\usepackage{xcolor}
\definecolor{codebg}{gray}{0.96}
\usepackage[round,authoryear]{natbib}
\usepackage[hidelinks,breaklinks=true]{hyperref}
\usepackage[capitalize]{cleveref}

\title{TRACE: A taxonomy-grounded synthetic dataset for teaching-program generation and session interpretation in Applied Behavior Analysis}

\author{Festus Kahunla\\
Drexel University \textperiodcentered{} Pombo Labs\\
\texttt{fmk39@drexel.edu}}

\date{}

\begin{document}
\maketitle

\begin{center}
\small
Code repository: \url{https://github.com/Pombo-Labs/TRACE}\\[2pt]
Dataset: \url{https://huggingface.co/datasets/PomboLabs/TRACE}
\end{center}

% ======================================================================
\begin{abstract}
\noindent
Applied Behavior Analysis (ABA) is a clinical discipline whose documentation, teaching programs and multi-session behavioral logs, is formulaic and high-volume, yet real session data is HIPAA-protected and bound by professional confidentiality rules, blocking the release of a training corpus. We present \textbf{TRACE} (\textbf{T}axonomy-\textbf{R}eferenced \textbf{A}BA \textbf{C}linical \textbf{E}xamples), a 2{,}999-example synthetic instruction-tuning dataset covering two ABA tasks: teaching-program generation across Discrete Trial Training, Natural Environment Teaching, and Task Analysis; and multi-session behavioral interpretation across twelve trajectory patterns and thirteen target behaviors. Every example is produced by a deterministic taxonomy-driven generator grounded in the canonical ABA literature, and every example carries complete sampling provenance, the exact taxonomy cells that produced it. The dataset is released under CC BY-NC~4.0 for data and MIT for code, with stratified train (2{,}549), validation (149), test (281), and sanity (20) splits. TRACE is a research artifact and has not been clinically validated.
\end{abstract}

% ======================================================================
\section*{Background \& Summary}
% ======================================================================

Applied Behavior Analysis is the dominant evidence-based behavioral treatment for autism spectrum disorder in North America, with approximately 89{,}000 certified BCBAs and BCaBAs worldwide \citep{bacb2026data}. Two document types recur constantly in clinical practice. \emph{Teaching programs} are method-specific specifications for acquiring a discrete skill, discriminative-stimulus design, prompt hierarchy, reinforcement schedule, error-correction procedure, mastery criteria, generalization plan, rendered in a register that varies by teaching method. \emph{Session interpretations} are clinical summaries drawn from multi-session behavioral logs that report per-program accuracies alongside per-behavior measurements (frequency, duration, partial-interval recording, inter-observer agreement); the interpreter classifies a trajectory pattern, hypothesizes behavior functions, and produces programming recommendations and, when warranted, a crisis plan.

Building a language model that drafts these documents is gated by data access. Real session data is HIPAA-protected and bound by professional confidentiality rules \citep{bacb2020}; de-identified releases cannot reliably preserve the clinical detail required for training. \citet{peck2025chatgpt} reported that BCBAs in a blind comparison preferred ChatGPT responses to clinician responses on ABA questions, motivating the application space and simultaneously raising the stakes of hallucination. The ethical use of AI in ABA service delivery is itself an active question in the field \citep{jenningscox2024ethics}. The closest prior work \citep{kumar2024personalized} generates ABA treatment plans and skill-acquisition programs, trained on a proprietary provider dataset rather than a released corpus, and does not address multi-session interpretation. Generative AI in autism \citep{sohn2025genai} and small language models in healthcare \citep{garg2025slms} are active research areas.

TRACE is a synthetic instruction-tuning corpus covering both document types. It is generated from a controlled vocabulary whose every cell cites a source: the canonical ABA textbook \citep{cooperheronheward2020}; the functional-analysis and functional-behavior-assessment literature \citep{iwata1982/1994fa,hanley2003fba}; functional communication training \citep{carrdurand1985fct}; the Verbal Behavior Milestones Assessment and Placement Program (VB-MAPP) and Assessment of Functional Living Skills (AFLS) curricula; and the BACB Ethics Code \citep{bacb2020} together with the ABAI Position Statement on Restraint and Seclusion \citep{abai2010}. The controlled vocabulary is encoded as YAML, the generator is deterministic under \texttt{(configs, seed)}, and every example records the sampled cells in a \texttt{meta.provenance} field, making clinical audit and targeted repair straightforward.

TRACE is, to our knowledge, the first synthetic ABA corpus to cover both tasks. Clinical-accuracy refinement is performed at the taxonomy layer: a flagged example traces through \texttt{meta.provenance} to the responsible cell, a single edit resolves the class of examples affected, and the corpus regenerates deterministically, a property bootstrapped synthetic pipelines do not offer naturally. Only the taxonomy is ABA-specific; templates, compatibility rules, and generator code are domain-agnostic, so the pipeline transfers to other clinical disciplines with structured conventions. TRACE is released as a research artifact. It has not been clinically validated and is not a clinical tool; any use in a clinical setting is the user's and the facility's own responsibility.

% ======================================================================
\section*{Methods}
% ======================================================================

\subsection*{Taxonomy and clinical grounding}

The controlled vocabulary is distributed across YAML files organized by clinical area: \texttt{configs/shared/} for cross-area primitives (learner profiles, mastery states, prompt types); \texttt{configs/\{dtt,net,task\_analysis\}/} for the three teaching methods (each with a \texttt{taxonomy.yaml}, an assistant-output \texttt{template.yaml}, and a \texttt{compatibility.yaml} of clinical-plausibility constraints); and \texttt{configs/session\_interpretation/} which additionally carries \texttt{trajectory\_rules.yaml} (pattern-specific accuracy and behavior-frequency generators) and \texttt{recommendations.yaml} (per-pattern antecedent/replacement/consequence/crisis bullet pools).

Each taxonomy entry cites a source: \citet{cooperheronheward2020} for response-class operational definitions, NET, chaining, and the systematic-desensitization basis of the toleration variant; \citet{lovaas1987} and \citet{smith2001dtt} for DTT; \citet{iwata1982/1994fa} for functional-analysis methodology and \citet{hanley2003fba} for the consolidated four-function taxonomy; \citet{carrdurand1985fct} for FCT and replacement behavior; \citet{stokesbaer1977generalization} for generalization; \citet{touchettehoward1984errorless} for time-delay prompting; \citet{bacb2020} (section 3.05) and \citet{abai2010} for crisis-plan content. Crisis-plan content deliberately refrains from specifying restraint procedures, those vary by jurisdiction, training certification, and learner-specific contraindications. Every crisis-plan bullet embeds BIP-authorization language, de-escalation-first sequencing, and contraindication-aware defaults.

\subsection*{Generation loop}

Per example: (1) sample a clinically-valid configuration from the taxonomy, weighted by clinical frequency; (2) apply compatibility rules to reject inconsistent combinations (for example, \emph{errorless} error correction pairs only with \emph{most-to-least} prompting per \citealp[ch.~21]{cooperheronheward2020}); (3) compute template slots from the sampled cells; (4) render the \texttt{user} and \texttt{assistant} messages; (5) stamp gold labels and complete sampling provenance.

Formally, TRACE v1 is $\mathcal{D} = \{e_i\}_{i=1}^{N}$ with $N=2{,}999$, where each example is a deterministic function of its sampled taxonomy cells and a seed:
\begin{equation*}
e_i = \Phi_{a(i)}\!\left(c_i;\; T_{a(i)},\; \sigma_i\right), \qquad c_i \in \mathcal{C}_{a(i)}.
\end{equation*}
Here $a(i) \in \{\text{dtt},\text{net},\text{ta},\text{si}\}$ identifies the clinical area, $\mathcal{C}_{a(i)}$ is the compatibility-filtered taxonomy-cell space for that area, $T_{a(i)}$ is the area's assistant-output template, and $\sigma_i$ is the seed state consumed at example $i$. Gold labels are a projection $\pi(c_i) \subset c_i$ and the full provenance is $c_i$ itself, recorded in \texttt{meta.provenance.taxonomy\_cells}. Re-running the pipeline with the same initial seed reproduces $\mathcal{D}$ exactly.

Session logs are constructed in layers: a learner profile and 3--6 programs (a mix of acquisition targets and, where applicable, FCT-style replacement responses); per-program per-session accuracy trajectories driven by pattern-specific generators (\emph{mastery progression}, \emph{regression}, \emph{extinction burst}, \emph{setting event trigger}, etc.); 0--3 target behaviors with frequency trajectories; optional antecedent-behavior-consequence entries on approximately 30\% of sessions with behaviors; an inter-observer-agreement session on approximately 25\% of logs; and a pattern-matched behavioral-indicator cluster. Interpretation content (clinical concerns, pattern classification, behavior-function hypotheses, recommendations, crisis plan) is then generated from the same provenance state.

\subsection*{Behavior-specific measurement}

A generic frequency line loses clinically important structure for several behaviors; TRACE uses behavior-specific shapes (\cref{tab:measurement}).

\begin{table}[h]
\centering
\small
\begin{tabular}{p{3.5cm}p{9cm}}
\toprule
\textbf{Behavior} & \textbf{Measurement shape} \\
\midrule
Tantrum & \texttt{freq~N, duration~Mm~total} \\
Stereotypy, mouthing & \texttt{freq~N; PIR~P\%} \\
Pica & \texttt{attempts~N (X~unsuccessful, Y~successful)} \\
Fecal smearing (scatolia) & \texttt{attempts~N (X~intercepted, Y~completed)} \\
Toileting & \texttt{urine: X~in-toilet / Y~accidents; BM: P~in-toilet / Q~accidents} \\
\bottomrule
\end{tabular}
\caption{Behavior-specific measurement shapes in session-log data. Generic frequency is retained for other behaviors (aggression, SIB, elopement, property destruction, non-compliance, verbal aggression).}
\label{tab:measurement}
\end{table}

\subsection*{Refineability}

The dataset is refineable at the taxonomy layer: any flagged inaccuracy traces through \texttt{meta.provenance} to the responsible cell, and editing that cell plus regenerating the corpus systematically updates every example that sampled it. No per-example rewrites are required, and the effect of each small edit is systematic across the full corpus.

% ======================================================================
\section*{Data Records}
% ======================================================================

Each example is one JSONL line carrying a chat triple and a \texttt{meta} block:

\begin{lstlisting}
{"messages": [
   {"role": "system",    "content": "<ABA clinical-assistant system prompt>"},
   {"role": "user",      "content": "<task prompt>"},
   {"role": "assistant", "content": "<structured clinical response>"}],
 "meta": {"task_type": "teaching_program" | "session_interpretation",
          "example_id": "<16-hex deterministic content hash>",
          "gold_labels": {...},
          "provenance": {"area": "...", "template_id": "...",
                         "taxonomy_cells": {...},
                         "seed_tag": "...", "generated_at": "..."}}}
\end{lstlisting}

The \texttt{example\_id} is a SHA-256 hash of the message content, truncated to 16 hex characters.

\paragraph{Gold labels.} For teaching programs: \texttt{method}, \texttt{domain}, \texttt{level} (VB-MAPP only), \texttt{learner\_profile}, \texttt{mastery\_state}, plus \texttt{program\_type} and \texttt{chain\_type} for Task Analysis. For session interpretation: \texttt{pattern\_class} (one of twelve), per-behavior \texttt{behavior\_functions} in the standard four-function taxonomy, an ordinal \texttt{escalation\_level} (1--4), a \texttt{confidence} level (high/moderate/low), and a Boolean \texttt{crisis\_plan\_required}.

\paragraph{Splits.} The repository ships four JSONL files (\cref{tab:splits}), stratified by area~$\times$~category so each split mirrors the corpus distribution. The test split is the held-out curation pool minus a 20-example stratified sanity carve-out.

\begin{table}[h]
\centering
\small
\begin{tabular}{lrrl}
\toprule
\textbf{File} & \textbf{Examples} & \textbf{Fraction} & \textbf{Purpose} \\
\midrule
\texttt{train.jsonl}  & 2{,}549 & 85.0\% & Fine-tuning \\
\texttt{valid.jsonl}  &   149   &  5.0\% & Validation during training \\
\texttt{test.jsonl}   &   281   &  9.4\% & Held-out evaluation \\
\texttt{sanity.jsonl} &    20   &  0.7\% & Training smoke-test \\
\midrule
\textbf{Total}        & 2{,}999 & 100\%  & \\
\bottomrule
\end{tabular}
\caption{File records and split composition. Stratification key: area~$\times$~category.}
\label{tab:splits}
\end{table}

\paragraph{Corpus composition.} The 2{,}999-example corpus decomposes across four areas: 800 Discrete Trial Training, 500 Natural Environment Teaching, and 499 Task Analysis examples (425 independence and 74 toleration) for the teaching-program task, and 1{,}200 multi-session logs for the interpretation task. The session-interpretation partition is spread approximately uniformly across twelve trajectory patterns, with per-pattern counts ranging from 84 to 128 examples.

\paragraph{Learner-profile distribution.} Four developmentally-anchored learner profiles span the corpus (\cref{tab:profiles}). The profiles are age-aligned to the curricula they draw from: \emph{early} learners map to VB-MAPP Levels 1--2, \emph{school-age} to VB-MAPP L2--L3 and AFLS basic-living / home-skills, \emph{adolescents} to AFLS community and vocational modules, and \emph{adults} primarily to AFLS independent-living.

\begin{table}[h]
\centering
\small
\begin{tabular}{lrr}
\toprule
\textbf{Learner profile} & \textbf{Count} & \textbf{Share} \\
\midrule
Early learner (approx.\ 3--5 yrs)        & 1{,}077 & 35.9\% \\
School-age (approx.\ 6--10 yrs)          &   747   & 24.9\% \\
Adolescent (approx.\ 11--17 yrs)         &   622   & 20.7\% \\
Adult (approx.\ 18+ yrs)                 &   553   & 18.4\% \\
\midrule
\textbf{Total}                           & 2{,}999 & 100\%  \\
\bottomrule
\end{tabular}
\caption{Learner-profile distribution. Developmental ages are approximate descriptors; the dataset does not encode clinical diagnosis, race, ethnicity, socioeconomic status, or gender.}
\label{tab:profiles}
\end{table}

\paragraph{Taxonomy feature coverage.} The generator samples from a controlled vocabulary with the following cardinalities (values given where compact; full enumerations and citations are in the taxonomy reference file \texttt{docs/taxonomy-v1.md}). \emph{Teaching-program dimensions:} 3 teaching methods (DTT, NET, Task Analysis), 3 VB-MAPP levels (L1--L3), 5 AFLS modules, 6 mastery states (emerging, developing, approaching, near, mastered, generalization), 2 program types for Task Analysis (independence, toleration), 3 chain types (forward, backward, total-task), 6 prompt hierarchies (most-to-least, least-to-most, time delay, graduated guidance, stimulus fading, stimulus shaping), 7 reinforcement schedules (CRF, FR-2, VR-3, token economy, CRF-per-step, terminal, differential-per-step), 8 error-correction procedures, 19 mastery criteria, and 182 distinct skill targets drawn from VB-MAPP and AFLS skill lists. \emph{Session-interpretation dimensions:} 12 trajectory patterns (mastery progression, regression, plateau, frustration, variable performance, prompt dependency, rapid acquisition, generalization failure, extinction burst, skill loss after break, motivating-operation shift, setting-event trigger), 13 target behaviors (tantrum, aggression, SIB, elopement, property destruction, motor stereotypy, vocal stereotypy, non-compliance, mouthing, pica, verbal aggression, fecal smearing, toileting accidents), 4 behavior functions (escape, attention, tangible, automatic), a 4-level ordinal escalation label, and 3 confidence levels. Compound cells add further combinatorial breadth: \texttt{behavior\_ids} realizes 86 distinct multi-behavior subsets across logs, and \texttt{behavior\_functions} realizes 14 distinct per-log function multisets. Synthetic learner identifiers follow a \texttt{SYN-\#\#\#\#} pattern from a fixed range; synthetic dates fall within 2026-01-01 to 2026-12-31.

\paragraph{Hosting.} The dataset is hosted on the Hugging Face Hub at \url{https://huggingface.co/datasets/PomboLabs/TRACE} and mirrored in the GitHub repository at \url{https://github.com/Pombo-Labs/TRACE}. The Hugging Face record ships with a dataset card, a Gebru datasheet \citep{gebru2021datasheets}, and a Bender-Friedman data statement \citep{benderfriedman2018datastatements}.

% ======================================================================
\section*{Technical Validation}
% ======================================================================

TRACE validation covers five mechanically-checkable properties plus a clinical-review pass.

\textbf{Schema integrity.} All 2{,}999 released examples parse as JSON, carry a three-message chat triple (system, user, assistant), and expose the complete \texttt{meta} block described in \emph{Data Records}.

\textbf{Provenance integrity.} 100\% of examples (2{,}999/2{,}999) carry a populated \texttt{meta.provenance.taxonomy\_cells} field. Across the corpus, taxonomy cells span twenty dimensions: for teaching programs these include \texttt{skill\_target} (182 unique targets), \texttt{prompt\_hierarchy} (6 values), \texttt{reinforcement\_schedule} (7 values), \texttt{error\_correction} (8 values), \texttt{mastery\_criterion} (19 values), \texttt{module} (5 AFLS modules), \texttt{chain\_type} (3 values), \texttt{mo\_arrangement} (9 values), \texttt{mo\_category} (15 values), \texttt{prompt\_strategy} (7 values), and \texttt{shaping\_n\_steps} (3 values). For session interpretation the provenance fields include \texttt{hidden\_pattern\_id} (12 patterns), \texttt{learner\_profile} (4 profiles), \texttt{n\_sessions} (8 distinct log lengths), \texttt{n\_programs} (4 values), \texttt{n\_behaviors} (3 values), \texttt{behavior\_ids} (86 distinct behavior subsets), \texttt{behavior\_functions} (14 distinct per-log function multisets), \texttt{has\_abc\_data}, and \texttt{has\_ioa\_session}. Gold labels are derived deterministically from the sampled cells, so labels are correct-by-construction relative to the taxonomy (label noise reduces to taxonomy-definition noise rather than annotation noise).

\textbf{Uniqueness.} All released \texttt{example\_id} values are unique within and across splits.

\textbf{Stratification.} Splits are produced by a largest-remainder stratified sampler over the category axis (method for teaching programs, pattern\_class for session interpretation). Per-pattern counts in the test split range from 8 to 12 examples across the twelve patterns, proportional to the full corpus to within two examples per pattern. The test split contains 75 DTT, 47 NET, 47 Task Analysis, and 112 session-interpretation examples.

\textbf{Determinism.} Running \texttt{src/generate.py -{}-all} twice with the same \texttt{seed} (set in \texttt{configs/generation.yaml}) produces identical corpora: identical \texttt{example\_id} hashes, identical split contents. Determinism is enforced by routing every random draw through a single seeded \texttt{random.Random} instance per area.

\textbf{Clinical review.} Clinical-accuracy review was conducted against published sources. For multi-reviewer BCBA validation, the recommended scoring is quadratic-weighted Cohen's $\kappa$ \citep{cohen1968weightedkappa} on the ordinal escalation-level head and the Matthews correlation coefficient with macro-F1 \citep{chiccojurman2020mcc} on the pattern-class head, against the four-dimensional rubric of \citet{nazar2025jmir}.

% ======================================================================
\section*{Usage Notes}
% ======================================================================

\paragraph{Loading.} The dataset loads with the Hugging Face \texttt{datasets} library:

\begin{lstlisting}
from datasets import load_dataset
ds = load_dataset("PomboLabs/TRACE")
ex = ds["train"][0]
print(ex["meta"]["gold_labels"])      # evaluation labels
print(ex["meta"]["provenance"])       # sampled taxonomy cells
\end{lstlisting}

Alternately, each split is a standalone JSONL file that can be iterated directly.

\paragraph{Reproducibility.} The corpus regenerates end-to-end from \texttt{(configs, seed)}:
\begin{lstlisting}
uv run python src/generate.py --all      # 3,000 examples across four areas
uv run python src/split_data.py          # deduplicate + stratified split
uv run python src/compile_curation.py    # test.jsonl + sanity.jsonl
\end{lstlisting}

\paragraph{Intended use.} (i) Supervised fine-tuning of small language models on ABA-flavored instruction-following with QLoRA \citep{dettmers2023qlora}; precedents include the on-device clinical LM Menta \citep{zhang2025menta} and MedGemma \citep{medgemma2025}, a medical adaptation of Gemma 3. (ii) Research into taxonomy-driven synthetic data generation with provenance. (iii) Baseline benchmarking for future ABA-specific language models.

\paragraph{Not for.} Autonomous clinical decisions. Training on or combining with real client records. Writing final Behavior Intervention Plans without clinician review. Medical diagnosis, legal documentation, or insurance reimbursement.

\paragraph{Evaluation guidance for reusers.} A recommended evaluation protocol for models trained on TRACE: LLM-as-judge rubrics adapted from Med-PaLM~2 \citep{singhal2025medpalm2}, HealthBench \citep{arora2025healthbench}, and TN-Eval \citep{shah2025tneval}, with Prometheus~2 \citep{kim2024prometheus2} as an open judge and judge-bias controls from \citet{zheng2023mtbench}; SelfCheckGPT \citep{manakul2023selfcheckgpt} for zero-resource hallucination probing; Expected Calibration Error \citep{guo2017calibration} for the confidence head. BLEU is unreliable as a primary metric \citep{reiter2018bleu}. Methodology precedent for synthetic clinical instruction-tuning data includes \citet{wang2023selfinstruct}, \citet{zhang2023alpacare}, and \citet{nazar2025jmir}.

\paragraph{Caveats.} Pattern frequencies across the twelve session-interpretation classes are approximately uniform for learnability rather than epidemiologically weighted; real clinical caseloads are skewed toward \emph{mastery progression}, and users training models with epidemiological validity should reweight. Sampling weights on other taxonomy dimensions are human-specified approximations where clinical frequency is not published in the underlying sources. The corpus is English-only and US-clinical in register, with VB-MAPP and AFLS curricula and BACB/ABAI ethics anchors. Teaching-method coverage in v1 is DTT, NET, and Task Analysis (independence and toleration); the Functional Communication Training \citep{tigerhanleybruzek2008fct}, Pivotal Response Training, and Behavioral Skills Training generators are architected and help design the taxonomy. The Behavioral Skills Training generator applies the instructions, modeling, rehearsal, and feedback protocol from \citet{parsonsrollysonreid2012bst} to teach learner skills. Session logs describe challenging behavior in the operational-definition register used in peer-reviewed ABA literature and may be distressing to readers unfamiliar with the field; no real person is described.

% ======================================================================
\section*{Code Availability}
% ======================================================================

All code used to generate TRACE is open-source and available at \url{https://github.com/Pombo-Labs/TRACE} under the MIT license. The generator (\texttt{src/generate.py} plus per-area generators in \texttt{src/generators/}) is a deterministic function of the taxonomy YAML files under \texttt{configs/} and a random seed specified in \texttt{configs/generation.yaml}; the v1 corpus regenerates end-to-end from source with one command. Python~3.10 or later and PyYAML are the only runtime dependencies. Scripts for stratified splitting (\texttt{src/split\_data.py}, which also deduplicates) and for rendering the held-out pool as a reviewable Markdown document (\texttt{src/prepare\_curation.py}) ship alongside the generator.

% ======================================================================
\bibliography{references}
% ======================================================================

% ======================================================================
% Tail statements per Scientific Data template
% ======================================================================

\section*{Data Availability}
The full TRACE v1 dataset is openly available on the Hugging Face Hub at \url{https://huggingface.co/datasets/PomboLabs/TRACE} and mirrored at \url{https://github.com/Pombo-Labs/TRACE} under CC BY-NC~4.0. No access restrictions apply for research use.

\section*{Ethics Declarations}
No human-participant data were collected. The dataset is entirely synthetic: no real client records, no real session notes, no real identifiers at any stage. No institutional review board approval was required. Synthetic learner identifiers use a \texttt{SYN-\#\#\#\#} pattern from a fixed range; synthetic dates fall within the 2026 calendar year. The dataset references crisis-intervention frameworks and restraint procedures only as examples; detailed restraint procedures are deliberately unspecified because they vary by jurisdiction, staff training certification, and learner contraindications.

\end{document}